\documentclass[sigconf]{acmart}

\usepackage{booktabs} % For formal tables

\setcopyright{rightsretained}
\acmDOI{10.475/123_4}

\acmISBN{123-4567-24-567/08/06}

\acmConference[2020]{arxiv}{February , 2020}{Dresden, Germany}

\acmArticle{4}
\acmPrice{15.00}

\begin{document}
\title{ Repository for Reusing Artifacts of Artificial Neural Networks }

%titlenote{http://Github.com/ghofrani85/RAN2/}

\author{Javad Ghofrani}

\orcid{0000-0002-9249-7434}
\affiliation{%
  \institution{Faculty of Informatics/Mathematics\\HTW University of Applied Sciences}
  %\streetaddress{Welfengarten 1}
  \city{Dresden}
  \state{Germany}
  \postcode{30167}
}
\email{javad.ghofrani@gmail.com}

\author{Ehsan Kozegar}
\orcid{0000-0002-1575-0651}
\affiliation{%
  \institution{Faculty of Engineering (Eastern Guilan) \\ University of Guilan
 }
  %\streetaddress{Welfengarten 1}
  \city{Guilan}
  \state{Iran}
 % \postcode{30167}
}
\email{kozegar@guilan.ac.ir}

\author{Hongfei Chen}
\affiliation{%
 \institution{Pitzer College}
 \streetaddress{}
 \city{Claremont}
 \state{CA, USA}
 \postcode{91711}
}
\email{hochen1998@gmail.com}

\author{Arezoo Bozorgmehr}
\affiliation{%
  \institution{Institute of General Practice and Family Medicine\\ University Hospital Bonn}
  %\streetaddress{Welfengarten 1}
  \city{Bonn}
  \state{Germany}
 % \postcode{30167}
}
\email{arezoo.bozorgmehr@ukbonn.de}

\author{Mohammad Divband Soorati}
\orcid{0000-0001-6954-1284}
\affiliation{%
  \institution{School of Electronics and Computer Science, University of Southampton }
  %\streetaddress{Welfengarten 1}
  \city{Southampton}
  \state{UK}
 % \postcode{30167}
}
\email{m.divband-soorati@soton.ac.uk}
 
\author{Dirk Reichelt}
\affiliation{%
  \institution{Faculty of Informatics/Mathematics\\HTW University of Applied Sciences}
  %\streetaddress{Welfengarten 1}
  \city{Dresden}
  \state{Germany}
 % \postcode{30167}
}
\email{dirk.reichelt@htw-dresden.de}

\author{Maximilian Naake}
\affiliation{%
  \institution{Faculty of Informatics/Mathematics\\HTW University of Applied Sciences}
  %\streetaddress{Welfengarten 1}
  \city{Dresden}
  \state{Germany}
 % \postcode{30167}
}
\email{maximilian.naake2@htw-dresden.de}

\renewcommand{\shortauthors}{J. Ghofrani et al.}

%synergy of AI and SE
\begin{abstract}
 %Artificial neural networks, especially deep neural networks, have shown many advantages and advances towards conventional software systems in  various fields such as machine translation, natural language processing and image processing. However, software engineering concepts, especially reuse,are not effectively investigated in this domain. In this regard, this paper proposes an online tool to support the re-usability in the development of artificial neural networks. Main contribution of our paper is to provide a cooperation platform between machine learning experts  and enable them to reuse the  artifacts used to develop ANNs. This tool is a single purposed repository---focusing on ML tools and their implementations---provides a rich dataset of artifacts that enables SPL experts to investigate various aspects of  reusability in the DNN as an emerging technology.

%final
Artificial Neural Networks (ANNs) replaced conventional software systems in various domains such as machine translation, natural language processing, and image processing. So, why do we need an repository for artificial neural networks? 
Those systems are developed with labeled data and we have strong dependencies between the data that is used for training and testing our network. Another challenge is the data quality as well as reuse-ability. There we are trying to apply concepts from classic software engineering that is not limited to the model, while data and code haven't been dealt with mostly in other projects.
The first question that comes to mind might be, why don't we use GitHub, a well known widely spread tool for reuse, for our issue. And the reason why is that GitHub, although very good in its class is not developed for machine learning appliances and focuses more on software reuse. In addition to that GitHub does not allow to execute the code directly on the platform which would be very convenient for collaborative work on one project.
%have not received enough attention in the literature so far.
%an area that warrants further work to move toward
%semi_final
We propose an online single-purpose repository of artifacts used during the development of ANNs as a cooperation platform for machine learning experts. 
Users can search, share, or reuse an existing artifact and submit their reviews to promote high value artifacts.
Given the costs of developing ANNs---large datasets and high computational power---our tool facilitates reusability and therefore, saves time and resources. 
We aim at providing an infrastructure for a thorough investigation of reusability in the domain of ANNs.  

%ongoing
%- benefits gained from this tool

%it provides a platform that facilitites investiation of  
%various aspects of reusablity in the domain of ann

%trash
%The tool wi a single purposed repository---focusing on ML tools and their implementations---provides a rich dataset of artifacts that enables experts of software engineering to investigate various aspects of reusability in the DNN as an emerging technology

%that facilitates reusing artifacts.

%- features of the tool?
%users can share their artifacts. they can search and  reuse the existing artifacts in partial or holistic form. furthermore, they can submit some feedbacks as review/ ranking that improves the reusablityil for the other users

%Our tool introduces a repository that focuses on providing a rich dataset of artifacts that allows software engineers to  investigate various aspects of reusability in the ANN through this platform
\end{abstract}




\maketitle

%\input{samplebody-conf}

%maximum of 6 pages!

\section{Introduction}

%\textcolor{red}{tracablity of artifacts}
%\textcolor{red}{interdciplinary work is more possible}
%\textcolor{red}{references}
%\textcolor{red}{dont forget to upload the project to github}
Application of Artificial Neural Networks (ANN) is expanding dramatically~\cite{basheer2000artificial}. ANNs are considered as replacements for software systems in many domains such as image, language, and text processing. %Some of the applications are assistant systems in cars, smart home equipment, as well as robotics. 
Fast growth in  applying  Machine Learning (ML) based systems in futuristic projects like industrial internet of things, self-driving cars, and medicine shows the importance of these techniques and their impact on the future of research and development. Unlike conventional software systems that are based on intensive models, codes, and test cases, ANN based systems are developed using labeled data. Although recent improvements in technologies such as GPU clusters and Big Data analytic methods make it feasible to develop reliable systems using ANN, these methods still depend heavily to the data that are provided for training and testing such systems. Providing  high quality labeled data and tagged information that satisfy the requirements of developing ML based systems is challenging~\cite{amershi2019software,sculley2015hidden}. We tackle this problem by utilizing the well-known and established concept of `reuse'  in the development of conventional software systems~\cite{krueger1992software, clementsnorthrop2001}. Software reuse promises increasing costs while improving  time-to-market and increasing the quality of software products. Model based software development based on reuse is well studied, however, less attention has been made to applying  concept of reuse in the development of ANN based systems. Reusing artifacts of ANN based solutions is limited to models while data (test and training data) and code are less considered\cite{ghofrani2019reusability}. 
To a prior submission of this paper we received a lot of feedback concerning thoughts of additional features, issues others saw with the idea, points that needed to be explained in more detail as well as what makes our idea different from that of others. In this paper, we do not only introduce an online tool for reusing the data, structure, and the models that are generated and utilized through the development of ANN systems. We also want to address the feedback we have gotten. Furthermore we extend the concept of sharing and reusing the artifacts of ANN based software systems by providing the advantages of reusability. With this concept it is possible to reduce the times a solution for an issue has to be created which is often consuming time and resources. Users can share, create, and modify the artifacts formed in the projects. Additionally, they can reuse the projects from other uses by copying and extending them in their own work-space. Our proposed online tool enables experts of software engineering to have a closer look at various aspects of reusing ANNs. Furthermore, the provided functionalities of our tool enables the ANN developers to reduce their effort to provide artifacts from their projects.  

Section \ref{sec:tool} proposes our tool and explains its  main functionalities based on four use cases. In  Section \ref{sec:relatedwork}, we review current state of the practice regarding reuse in the context of development of ANN based systems. Section \ref{sec:conclusion} concludes  our paper and discusses the issues around our tool and possible future work.

%main functionalities--> 1) repository for sharing data and implemntation of machine learning based solutions 2) 
\section{Repository for Reusing Artifacts of ANN} \label{sec:tool}
\begin{figure}
    \centering
    \includegraphics[width=0.47\textwidth]{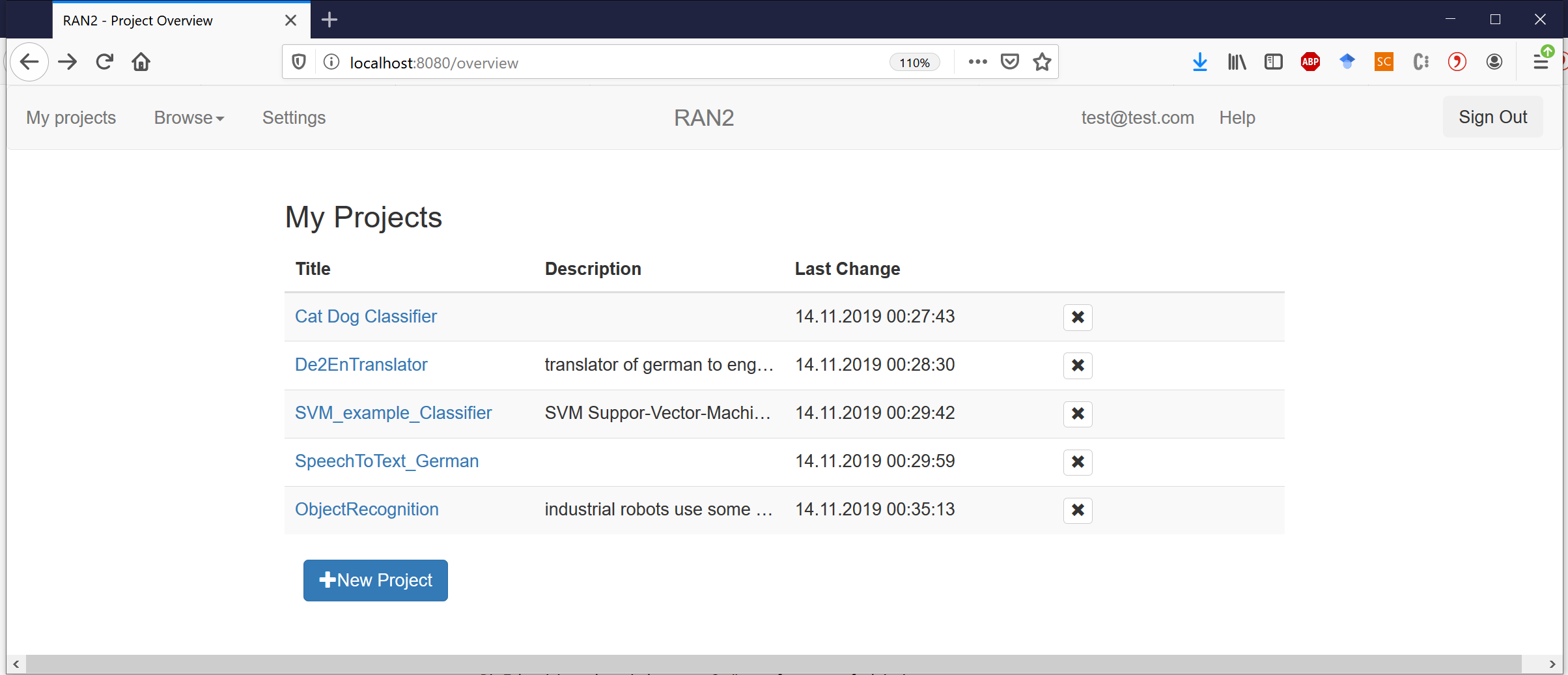}
    \caption{A view of user's Dashboard in RAN2}
    \label{fig:dashboard}
\end{figure}

In this section, we introduce the Repository for Reusing Artifacts of Artificial Neural Networks, shortly RAN2. Source code of the tool is publicly available in GitHub\footnote{\url{https://github.com/ghofrani85/RAN2.git}}.  %The basic functionality of user management that allows each user to register and receive an identity makes it possible to get authenticated and authorized for managing project within this tool. 
Users can register and receive an identity to be authorized for managing project within the tool. Users will be redirected to their home page after successful registration and login to RAN2. In the home page, users can create a new project or edit and view their existing projects listed on the page (see Figure \ref{fig:dashboard}). Users can also add labels/tags to their projects to allow other users to find the project using the search functionality.

\begin{figure}
    \centering
    \includegraphics[ width=0.5\textwidth]{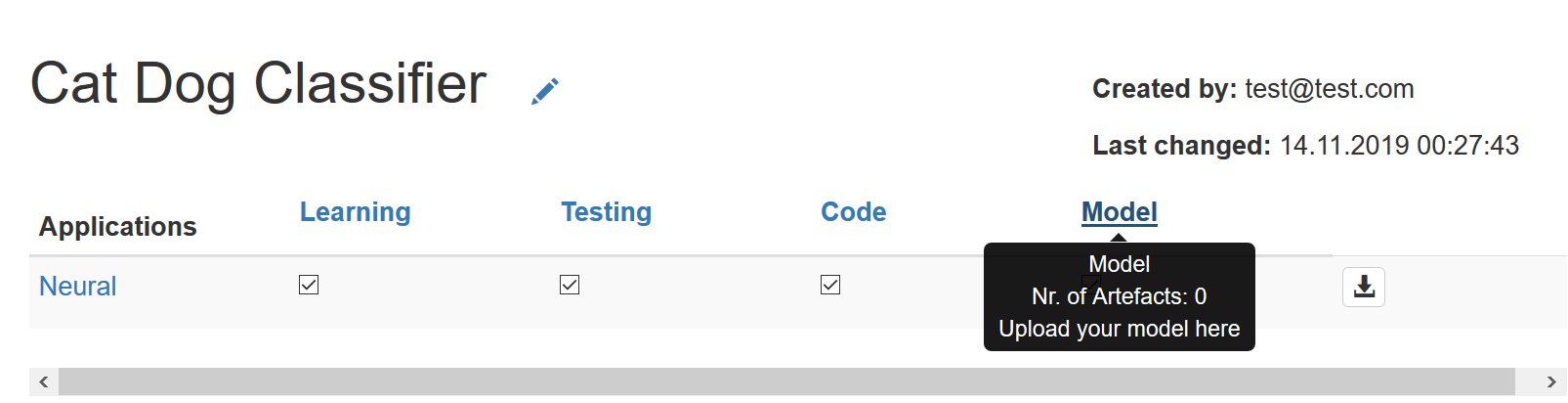}
    \caption{A view of four main folders under each project for storing related data to Test, Train, Model, Code in RAN2. Download button is in the right side. Selecting  each checkbox includes the contents of corresponding folders in the downloaded zip file that is generated by Download button}
    \label{fig:project1}
\end{figure}

Each project in RAN2 consists of four primary directories which are added to the project by default at the creation time. These folders contain four main categories from artifacts needed for developing ANN based solutions (see Figure~\ref{fig:project1}). The check-boxes under each folder allow users to select items to download a customized package. Users can click the \textit{Download} button to select a folder of the project and download it as a zip file. We decided on splitting the data for Training and Testing for a couple of reasons. The first and strongest is for the possibility to provide better comprehensibility. If another user can reproduce the model exactly as intended he can better understand what the original creator wanted to do. Secondly many times artificial data is used for training and then real data to test the outcome. Therefore a differentiation between these two is needed.

\begin{figure}
    \centering
    \includegraphics[width=0.48\textwidth]{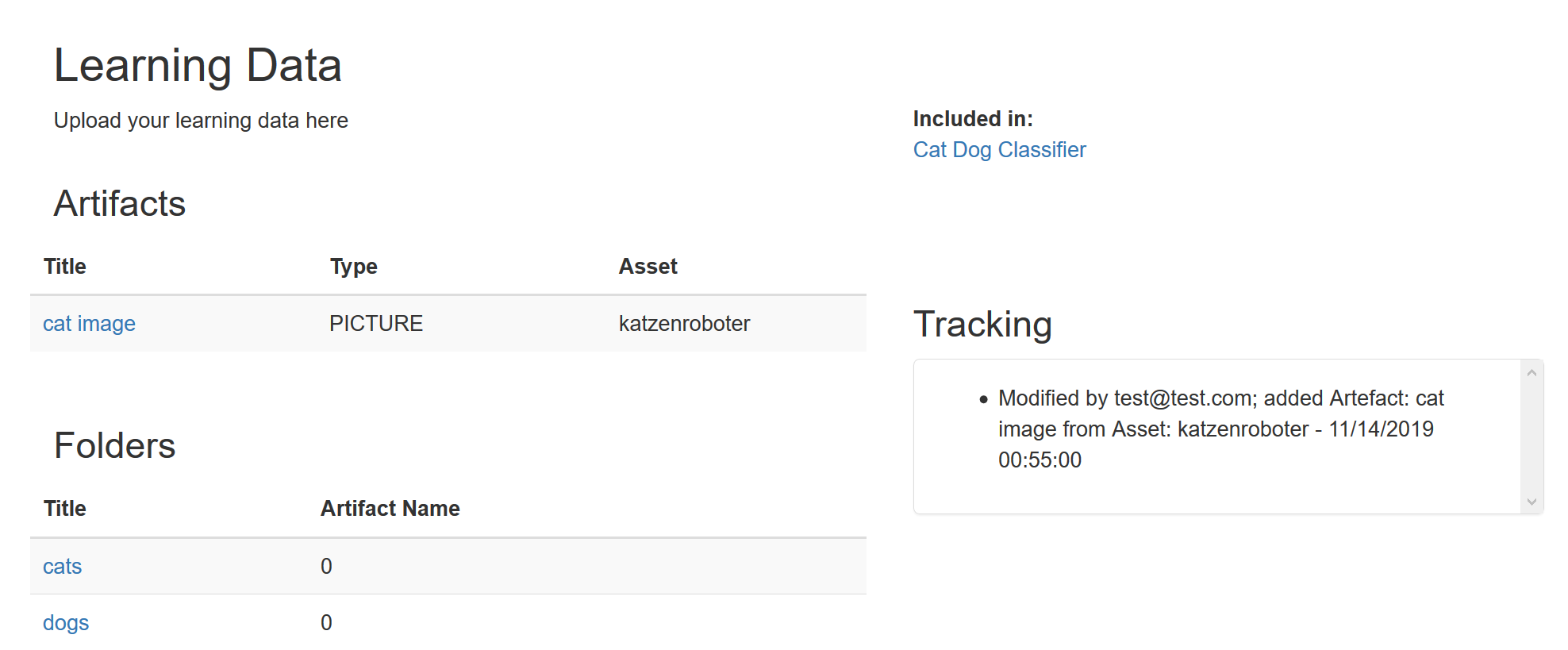}
    \caption{Example of contents of Learning Data Folder under one of projects in RAN2. The list of assigned artifacts are located in the  upper part of the page while the list of sub-folders are listed in the bottom of the page. Tracking window helps the user to follow the last changes that made by the users on the content of this folder}
    \label{fig:folderview}
\end{figure}

Furthermore, users can also preview the content of these folders by clicking on them. These folders may contain sub-folders or some artifacts. Figure~\ref{fig:folderview} illustrates a view on training data within a project. In this view, new sub-folders can be created, deleted, or renamed. New  artifacts can be added which are based on the underlying repository of reusable assets in RAN2. This repository is responsible for sharing and reusing the artifacts. This repository contains all assets that users provided for their projects. Within each folder view, users can add new artifacts by uploading an asset to the repository and select the whole or a part of it, or even reuse directly an existing asset from the repository.  An important feature is to label each asset with tags. Tags help finding and reusing the assets.  RAN2 supports users by embedded tools to modify pictures or parts of a video or audio files as well as text files. 

\begin{figure}
    \centering
    \includegraphics[width=0.47\textwidth]{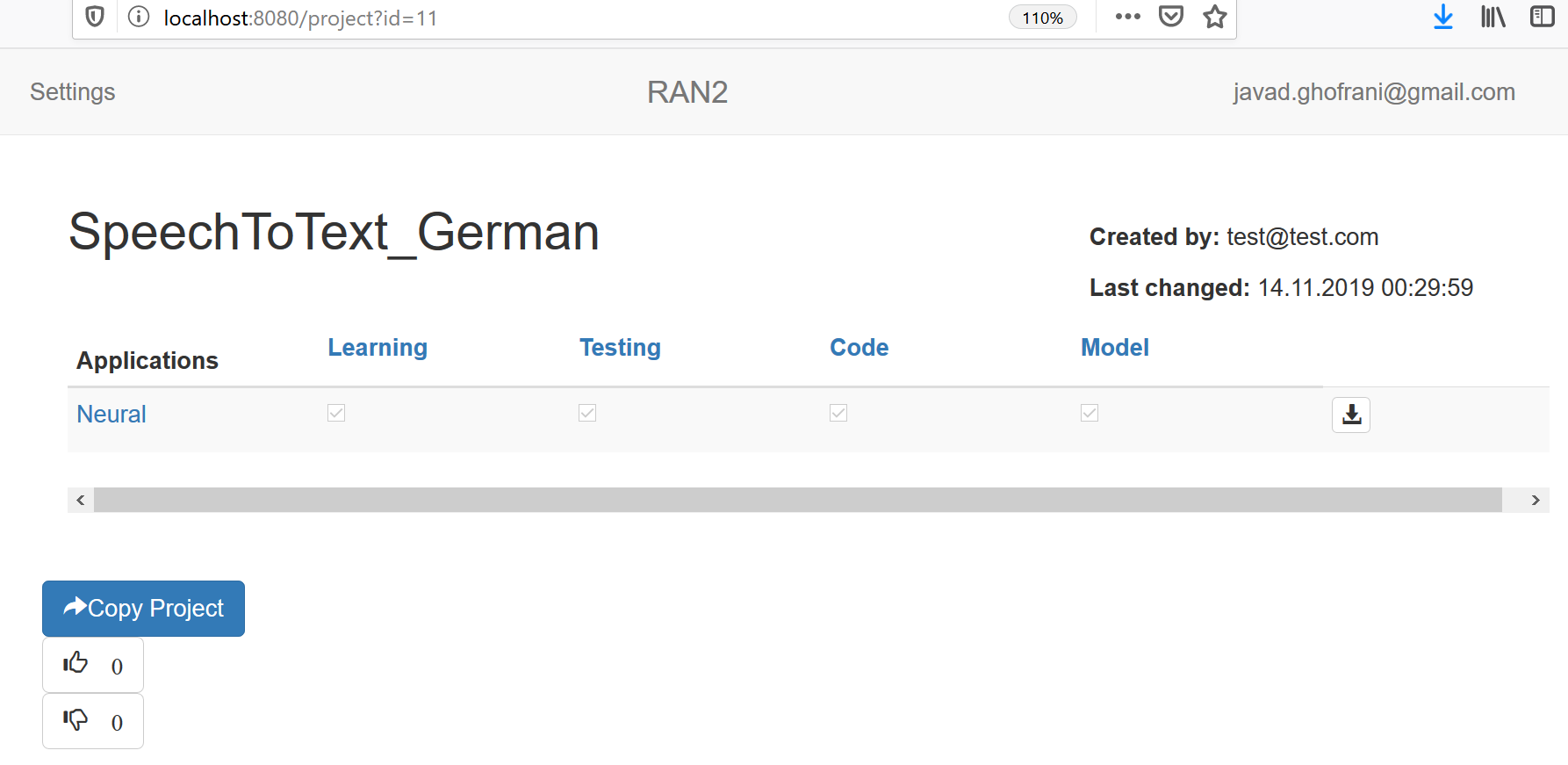}
    \caption{The projects from other users are available for the current user to copy. If the user copy an existing project from another user, two rate-up and rate-down buttons will appear to let the user give his/her feed-back about usefulness of the project. In this example the user javad.ghofrani@gmail.com has already copied a project from the user with identity of test@test.com. The user with identity of javad.ghofrani@gmail.com can rate-up or rate-down this project.}
    \label{fig:copyProject}
\end{figure}

\begin{figure}
    \centering
    \includegraphics[width=0.47\textwidth]{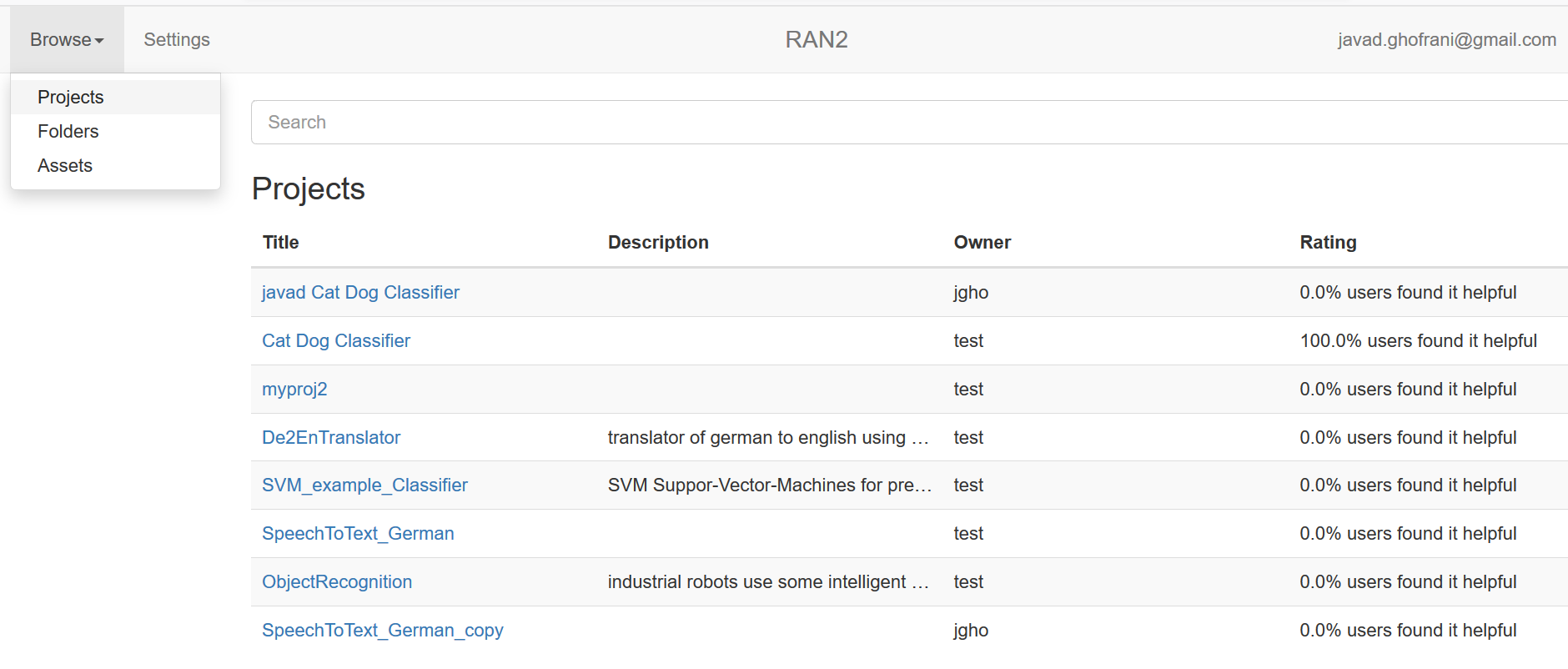}
    \caption{In the repository of all projects, the rating values for each project will be shown beside it}
    \label{fig:rating}
\end{figure}

Users can import the whole or parts of other projects into their project and customize it for their own needs. A copy button is available in the project that does not belong to the current user. After copying the projects, users can see a rate-up and rate-down button on the original project that allows them to give a feedback about the projects (see Figure~\ref{fig:copyProject}). Collected feedback from users will be shown beside the project in the repository view of projects in RAN2(see Figure~\ref{fig:rating}).

\begin{figure*}
    \centering
    \includegraphics[scale=0.70]{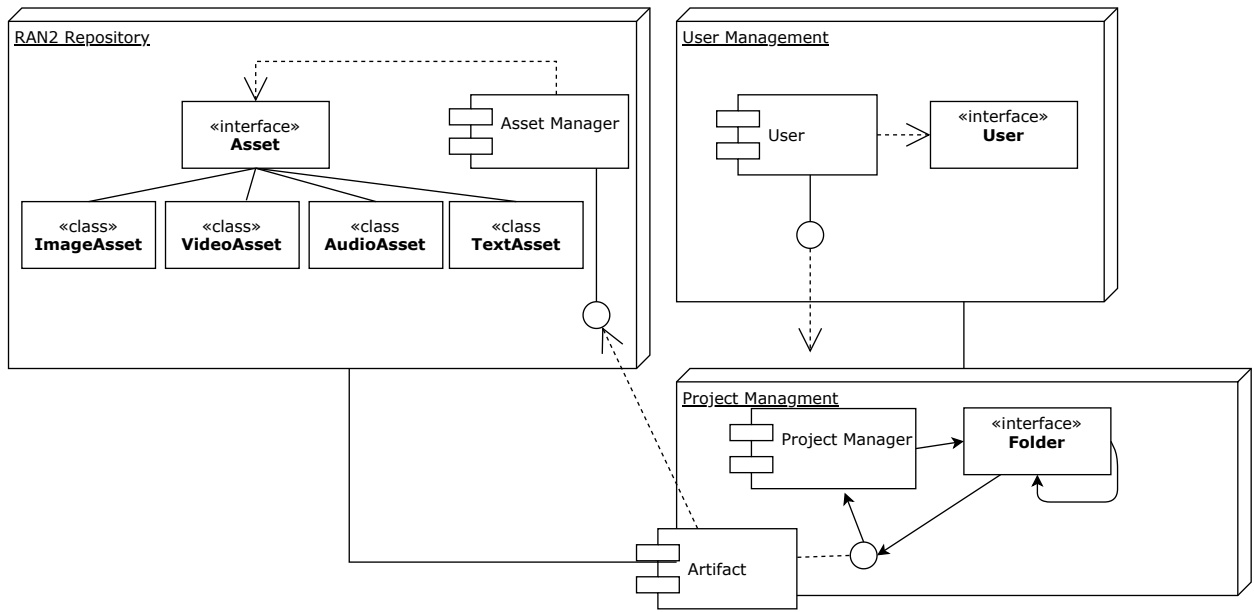}
    \caption{Component Diagram of RAN2 }
    \label{fig:workflowdiagram}
\end{figure*}

\subsection{Example Usage Scenarios}\label{sec:utilization}
 
%beside user management functionality such as sign in, sign up, and editing the user accounts, main functionality of our tool is managing the neural network projects. Through this functionality, the users can generate,edit(update) and remove their projects. one project can belong to only one user, however, the users can make fully or partial copies from other projects and make their own copies. Each project consists of input data in the form of train and test data which is created in the hierarchy of files in folders. We are planning the interfaces to bind databases or services which provide data and test artifacts for taring and testing a neural network.

%Furthermore, each project consists of a network which is provided by user (either hand-written code or generated using third party tools) or generated with the wizard of our tool. 
 %Reusing the projects in the whole or only some parts of the project is the main focus of our tool.
%For example, the user can only copy the data (both test and train data), or the network architecture (code) and use them  with a model from an online storage. 
%In order to make this functionality more clear, we introduce some examples of usage of our repository.
Some example scenarios were provided based on the proposed functionalities in RAN2. For development Java was used for the backend, Bootstrap for the frontend and all is based on a PostgreSQL database.

\begin{figure}[ht]
 \includegraphics[clip, trim=1cm 12cm 15cm 1cm,scale=0.6]{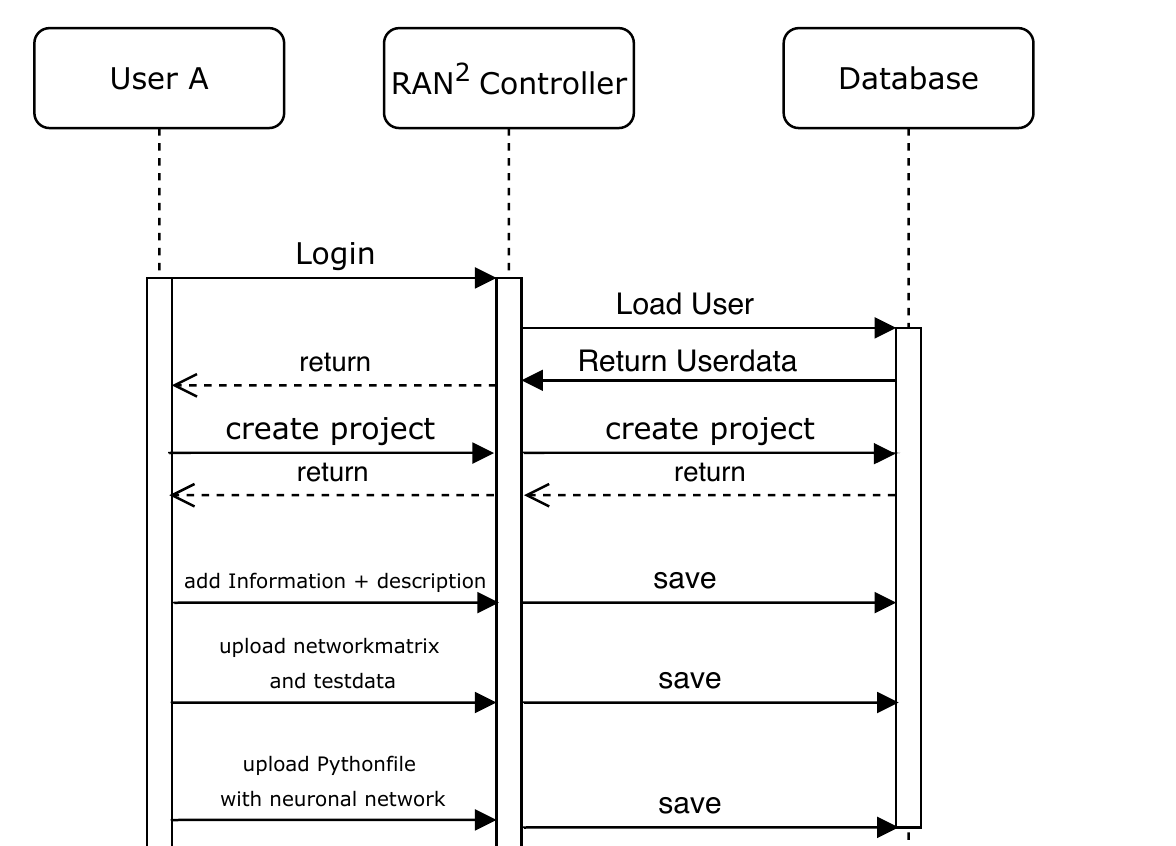}
 \caption{Sequence diagram example 1}
 \label{fig:usecase:1}
\end{figure}

\textit{Example 1. - figure \ref{fig:usecase:1}:} User $A$ sings up and creates a new project. She inserts some details about her network as description which helps the other users to find out what her network is aiming to do. She selects a $330\times330\times40$ network matrix and uploads her test and trains data including apple, orange, and pears. The user uploads a python file into the code category. This file contains a DNN developed using python and Keras. 

\textit{Example 2. - figure \ref{fig:usecase:2}:} User $B$ needs a network which can classify apple  images among a stream of images that he receives from a camera. He has developed a DNN in C++  to perform this task. He does not have enough input data (apple images) to train this DNN. He creates a new project and wants to add some training data. In this step, he searches in the existing projects of our repository using the keyword `Apple' and  finds the assets in the project  that have been created by user A. He selects the check boxes of the assets and copies the apple images (images with the apple label) into the category of train data in his new project. 

\begin{figure}[ht]
 \includegraphics[clip, trim=16cm 12cm 0cm 1cm,scale=0.6]{diagrams/Use-Cases.pdf}
 \caption{Sequence diagram example 2}
 \label{fig:usecase:2}
\end{figure}
\textit{Example 3. - figure \ref{fig:usecase:3}:} User $C$ needs a project for  classification of  apples, oranges, and carrots. She searches and finds the project of user $A$ that is generated with same objectives. In this case, user $C$ is not sure how good the quality and accuracy of the trained model and network in the project from $A$ is. Therefore, before starting to copy this project for herself, she downloads this project and performs some tests with some images from her use case. This way, she can see that whether the accuracy of that DNN satisfies needs of her project. 

\begin{figure}[ht]
 \includegraphics[clip, trim=1cm 1cm 15cm 12cm,scale=0.6]{diagrams/Use-Cases.pdf}
 \caption{Sequence diagram example 3}
 \label{fig:usecase:3}
\end{figure}

\textit{Example 4. - figure \ref{fig:usecase:4}:}  User $D$ needs a neural network which classifies oranges and apples in images. He finds the project from user $A$ by using search functionality of RAN2 and finds a similar project. He downloads a copy of this project and extends it by training this network with an additional set of data. This way, he improves the quality of the network with less training effort and time. Although this method is a common scenario among ANN developers (known as transfer learning), finding proper network is still challenging in such tasks.%After a while, user D needs to extend his network in the way, that the network is able to recognize the Pears beside other three fruits. He searches in the repository and find a dataset from a user who has some train and test data as a part of his project. he imports the data related to Pears into his project and trains it again
 
 \begin{figure}[ht]
 \includegraphics[clip, trim=16cm 1cm 0cm 12cm,scale=0.6]{diagrams/Use-Cases.pdf}
 \caption{Sequence diagram example 4}
 \label{fig:usecase:4}
\end{figure}

%It is clear that reuseing these artifacts reduces the time for providing data and  training the neural networks.

\section{Related Work}\label{sec:relatedwork}
This manuscript takes some of its inspirations from ProductLinRE\footnote{\url{http://www.productlinre.com}} introduced in our previous paper~\cite{ghofrani2018productlinre}. ProductLinRE is an online platform that enables the cooperate work on artifacts of Requirements Engineering (RE) in the development of conventional software systems. Using ProductLinRE, users can share and reuse  artifacts  such as text, images, video, and audio files that are involved in RE processes to reduce the effort and costs of generating new ones. However, these functionality is adapted for conventional software development without considering the ANN based solutions.    

% qualit / Standards in machine learning is the missing point (maybe here or in introduction as motivation)

%\textcolor{red}{kos ziad gofti inja be bad. dorostesh kon bad neda bede}

  Transfer learning \cite{torrey2010transfer} is an established method for reusing network structure and trained models of ANNs.  In this method, an existing pre-trained network is reused by extending its structure or retraining with a smaller set of data for customizing its functionality. This method saves time and computational resources in comparison to ANNs that are developed and trained from scratch.  Common datasets, such as Imagenet~\cite{deng2009imagenet} are used to provide pre-trained networks. The usual reused artifacts in transfer learning methods are network structure and models (weights of trained network).  Tools and frameworks such as Tensorflow\footnote{\url{https://www.tensorflow.org}} framework and Keras as deep learning library under Python support these way of reuse in the development of ANN based solutions. However, reusing the training sample is not supported in tools with transfer learning. Another disadvantage of transfer learning is the limited amount of well known datasets that contain pre-trained networks with visual data like images. 

%ModelZoo \footnote{\url{https://github.com/BVLC/caffe/wiki/Model-Zoo}} is a sharing folder on GitHub for models that are trained  for image classification tasks. This project is limited to the image classificaiton tasks while only models are supported. The other artifacts like test and train data are not supported in this work.  
%in this book, various simulation tools for neural networks are considered : 
%Teil III
%27.1 27.2 27.3 27.4 27.5 27.6 27.7 27.8
%Simulationstechnik Neuronaler Netze
%Kapitel 27 Software-Simulatoren neuronaler %Netze NeuralWorks Professional H/Plus
%BrainMaker
%Nestor Development System ANSimundANSpec %NEURO-Compiler
%NEUROtools SENN++
%Die PDP-Simulatoren
%27.9 RCS (Rochester Connectionist Simulator) %27.10 Neural Shell
%27.11 LVQ-PAKundSOM-PAK 27.12 Pygmalion
%27.13 SNNS (Stuttgarter Neuronale Netze %Simulator) 27.14 SESAME
%27.15 NeuroGraph 27.16 UCLA-SFINX
%27.17 PlaNet
%27.18 Aspirin/MIGRAINES 27.19 FAST
%27.20 VieNet2 27.21 Xerion
%27.22 GENESIS 27.23 MUME 27.24 MONNET 27.25 Galatea 27.26 ICSIM
%Kapitel 28 Der Stuttgarter Neuronale Netze Simulator (SNNS) Stuttgarter Neuronale Netze Simulator
%28.1
%28.1.1 Geschichte des SNNS 28.1.2 Struktur von SNNS
%28.1.3 Unterstützte Architekturen und Leistung Simulatorkern von SNNS
%28.2 28.3
%Graphikoberfläche von SNNS
%28.3.1 2D-Visualisierung von Netzen 28.3.2 Der Netzwerk-Editor
%28.3.3 3D-Netzwerk- Visualisierung 28.4
%Netzwerkbeschreibungssprache Nessus

%für weitere informationen hier gucken:https://modeling-languages.com/tools-modeling-artificial-intelligence-code/
Various online tools aim at facilitating the development of ANN based solutions without revealing their development effort. These tools are introduced from  open source foundations or commercial parties. We classified these tools based on their functionalities into two main categorized.  First category includes tools that provide visualization to enhance workflow definition and specify the input and output data to train a ANN. Examples for this category of tools are  RapidMiner\footnote{\url{https://rapidminer.com}} and Orange Data Mining Toolbox\footnote{\url{https://orange.biolab.si}}. Second category provides computing power as processing services or data-storage service to handle the complexity of training  ANNs.  Microsoft Azure Machine Learning Studio\footnote{\url{https://studio.azureml.net}} and IBM  SPSS Modeler~\footnote{\url{https://www.ibm.com/products/spss-modeler}} belong to this second category. The proposed work by Pahl and Loipfinger~\cite{pahl2018machine} follows the same strategy by providing encapsulated ML techniques as services which make it possible to reuse them in service-oriented architectures.
%\textcolor{red}{baz ke kos gofti: This difference originated from difference between methods that utilized for development of conventional software systems and ANN based systems.}
 The contribution of RAN2 compared to the ProductLinRE is the domain of usage.  ProductLinRE is developed to facilitate reusability in RE while RAN2 covers the entire development process of ANN based solutions. Note that sharing the resources and artifacts in RAN2 repository with the users of ProductLinRE, and vice versa is a possibility. 
%some of these tools are as following: , and a few . Main focus of these tools is to enhance the creating and the training of ML solutions with the help of web-based graphical user interfaces and data visualisation tools. Nevertheless, reusability of artifacts in such solutions are neglected in these solutions.  

%Pahl and Loipfinger~\cite{pahl2018machine} proposed REST ML services which could be bounded to web based project as a service. In this regard, they have  encapsulated three famous machine learning techniques \textcolor{red}{()} in the form of reusable microservices and separated their data modules to get use them as reusable services. However, this techniques requires new dataset each time that the users of such services have new or different requirements than those which are available. 

To end this section a comparison of RAN2 with other existing tools in this area will be provided. Tools RAN2 is competing with and where we're going to focus on are Google Colab, But4Reuse and OpenML.

Starting with Google Colab, it can be said that it offers a web application that allows a very direct collaboration on one model. The data can be shared selectively with other users. While it is certainly good that sharing with others is possible it is a bad thing that the sharing can not be done for all users of the platform and therefore there is no possibility to build a big community which shares and modifies their work all together. So Google Colab does not provide the functionality that RAN2 is trying to achieve.

The next tool is But4Reuse. It is one of the few tools that offers running the code in itself. This is great for really collaborative work because everyone can access the project and see the results. In addition to that it is very well documented so users can get into it easily. Not so good points of But4Reuse are that it supports only a limited amount of programming languages and that it is necessary to install an application, there is no web application for easier access.

OpenML is the most advanced tool in this comparison. It offers not only direct execution of flows and a pronounced tagging feature. It also has a big community of users that take part in the project. But OpenML could be criticized for not addressing the issue of data privacy, because everything is open to anybody on the platform. There is no function to make a project private. Additionally getting started is very hard and usage is complicated. 

When comparing RAN2 with the other tools, it can be seen that although it is still in an early stage of development and it does not offer direct running of machine learning processes, it offers a specific focus on artificial intelligence, in an easy to easy to start with and easy to use web application.

To conclude this section it can be said that there exists no tool that covers all problems in the field. Each addresses different issues and focuses on certain functionality, so each also has its own field of problems it faces. That is the chance for RAN2 to address these problems and to become a tool that solves these issues.

\section{Discussion}

 RAN2 is still in early stages of development. Some potential deficiencies and improper functionalities may degrade the quality of the tool and the user experience. However, the core features are functioning and the tool allows reusability in the development of  ANN based solutions in its current state. ANN based solutions often require numerous samples for training. An inevitable technical issue is providing a big storage, but we do not address this concern in the paper. As training sets are valuable assets in the development of ANN-based solutions, experts tend to protect and personalize it through copyrights. In RAN2, the users can upload images from other references, videos, or similar material, or even reuse it from other users. Therefore, copyright agreements have to be introduced to protect the rights of the users. 

Following the prior submission of this paper we had to categorize the feedback we got. There are four main points that can be identified: Additional Features, Distinguishing features, Explanatory issues of the paper and issues with the concept in general. While the Explanatory issues will be dealt with in their sections, in this section we are going to focus on a discussion of the other three main points. 

\subsection{Distinguishing features}
\begin{figure}[ht]
 \includegraphics[clip, trim=0.5cm 10cm 19cm 0cm,scale=0.7]{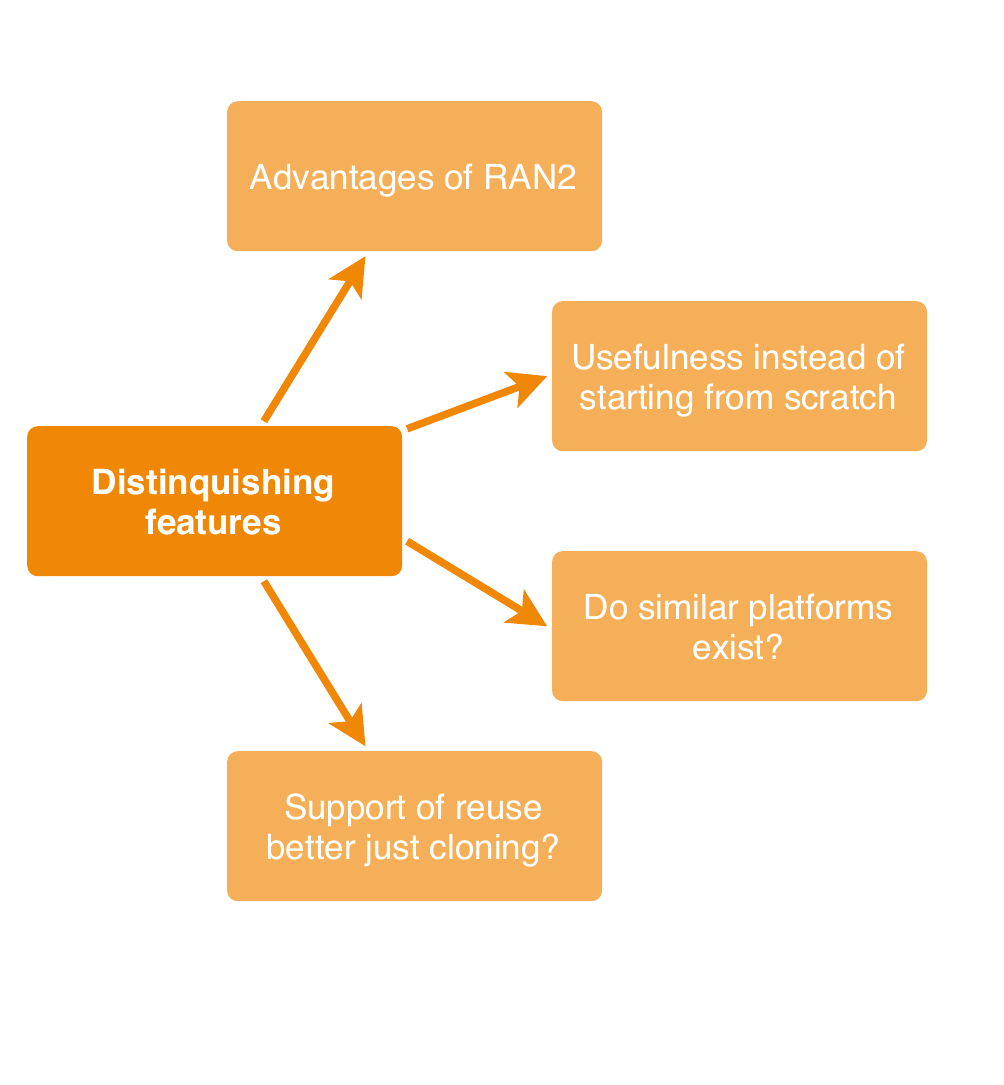}
 \caption{Distinguishing features}
 \label{fig:distfeat}
\end{figure}
As seen in figure \ref{fig:distfeat} these are features that set RAN2 apart from other tools that exist in this area. Mainly this is about which tools do exist and what advantages it provides compared to them. This was dealt with in the Related Work section of this paper.
Furthermore the question arose if it is useful anyways to reuse and not to start from scratch. To this point we can give a strong no as we are confident that developers of machine learning projects are able to save a lot of time especially when they are looking for fitting data to train their models with.
Lastly the point if it is better to reuse than to just clone. The beauty of RAN2 is, that both is possible. Developers can both just clone a project and modify it to their own needs and they are able to reuse them partly or completely fitting to their needs.

\subsection{Additional features}
\begin{figure}[ht]
 \includegraphics[clip, trim=10cm 10cm 1cm 0cm,scale=0.7]{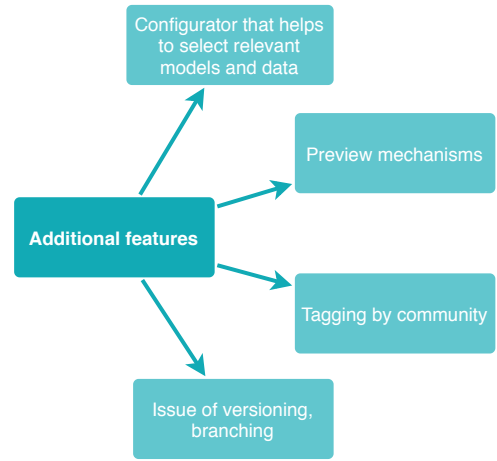}
 \caption{Additional features}
 \label{fig:addfeat}
\end{figure}
As mentioned before the users have certain needs and wants for a project like RAN2. The most wished for ones where put together in figure \ref{fig:addfeat}. All of those are good ideas where we should definitely think about how to incorporate them into RAN2. 

\subsection{Issues in general}
\begin{figure}[ht]
 \includegraphics[clip, trim=0.5cm 0cm 19cm 11cm,scale=0.7]{diagrams/feedback.pdf}
 \caption{Issues}
 \label{fig:iss}
\end{figure}
General issues others saw with the RAN2 project are displayed in figure \ref{fig:iss}. Those include mainly that the project is still in an early stage of development. In addition to that the tool will only gain usability when a critical number of users is engaging in it and participate in the reuse of models and data. This is am important point and it will show when users begin to use the platform. The last main point was about the confidentiality of data. For this issue we will build a functionality where users can decide how their data can be shared and reused. They can even make their project private. So the mentioned issues are not threatening to RAN2, but in further development adjustment need to be done.

%Another important issue in the development of ANN based solutions is  strong dependencies between the output accuracy and the quality of training samples. In order to reduce the effect of this fundamental issue in RAN2, we provide a review functionality, but these inputs are still delivered by users which may include biases.  by the     . The second challenge is the quality of projects and level of reusability that is fully depending on the provided inputs by  users. % Furthermore, we can utilize gamification to overcomellect points or stars based on their activities. It can motivate the users to maintain their projects and make it more reusable. % In order to prevent abuse of this system by the users, our tool should be able to perform some general test and create a bench-marking in certain periods of time and publish the quality of the projects based on their compatibilities with the benchmark inputs. 

   % Most of the time the developers of ANNs do not trust the code of others. It has the reason of being afraid of unknown source code with plenty of bugs or improper structure that makes it hard to reuse. However, our solution is not about reusing code written for ANNs. We are focusing on  reusing the data and share the data with other developers who need to reuse some labeled data or tags. We try to solve the problem of providing qualified  data for training purposes  which is more time consuming task in the development of ANNs. 

\section{Conclusion and Future work}\label{sec:conclusion}
In this paper, we introduced a web based tool for reusing the artifacts generated in the development of ANN based solutions. In order to offer a clear overview of these functionalities, we provided few examples of utilization that show how the usage of RAN2 can reduce costs (time) and efforts in the development of ANN-based solutions. RAN2 is still in early stages of development and future functionalities include (1) version controlling for creating branches and merging,  (2) improving the ranking system with textual reviews or comments, (3) enabling users to add additional information---meta data, description and documentation---to improve flexibility, transparency, and  reusability, and (4) providing a machine-to-machine interface for automated cooperation between systems to automatically search and find a solution and reuse it without any need for human interference. Other future work that is necessary is (5) how the moderation of the feedback feature should be realized and (6) what should be considered when the search feature is implemented. Other issues are (7) what happens when the origin of a cloned project is updated and (8) a further analysis of the needs of the end-user. 
 
    %We are going to implement a wizard that helps the user to create a new project in an step-by-step way. Using this wizard, the user can specify the structure of ANN and the type of  data (new or from other networks) to train the network. % Each project will enable the user to train and test the neural networks with the defined input and output.
%paragraph{d} automatic labeling, finding similarities and asking the user to add more information or suggest some information. \paragraph{d }private repositories which are not shareable with other users. 
%\paragraph{e} the user can create a new project with the help of a wizard. in this wizard the user can step by step select which type of network (matrix or vector), which type of the function (delta hepp), and which kind of data (new or from other networks) should be selected to train the network.  Each project enable the user to train and test the neural networks with the defined input and output. 

%\paragraph{e}Evaluation: measuring the reusability level for different types of project through different people with different levels of skills and knowledge. Usability 

%\paragraph{f} user can also perform some fine tuning or any changes in the structure of the neural network using a simple code editor.

%\section{Appendix}

%\end{document}
\bibliographystyle{ACM-Reference-Format}

\end{document}